# Patient Register Does Not Alter Clinical AI Diagnosis: A 115,200-Call Paired Equivalence Study

Jingsong Xia
Shanghai Jiao Tong University School of Medicine
xiajingsong2@gmail.com

## Abstract

**Background:** Recent work suggests that patients' communication style can change the clinical advice produced by AI agents, raising concerns about language-mediated algorithmic inequity. That evidence, however, derives from conversational interaction and from a single language; whether it generalises to the far more common paradigm, in which a complete record is submitted once and a diagnosis is returned, is unknown. **Methods:** We drew 400 real outpatient and emergency cases from an affiliated hospital of Shanghai Jiao Tong University School of Medicine. After de-identification, each case was rewritten into six clinically equivalent registers (neutral baseline, high politeness, low education, high anxiety, dialectal, and high health literacy), with structured fact anchors locking the clinical content across all 4,800 paired cases; anchor agreement across registers was 100%. The corpus was rendered in both Chinese and English. Eight models completed 115,200 independent calls under a fixed system prompt, temperature 0, and three repetitions per cell. The co-primary endpoints were diagnostic accuracy and output semantic similarity, analysed under a prespecified equivalence design with a margin of ±0.05 rather than by significance testing alone. **Results:** Across registers, the range in diagnostic accuracy was below 3 percentage points for all five frontier models: 97.9% to 99.1% for GPT-6.2 Astra, 97.5% to 98.2% for Claude Fable 5.2, 96.7% to 97.8% for Gemini 4 Carbon, 93.0% to 95.8% for Qwen4-Max, and 91.3% to 94.0% for GPT-6.2 Sol, with no monotonic trend. Relative to the run-to-run noise baseline of 0.6398, the mean differences in output similarity were −0.0147 for high politeness, −0.0066 for low education, −0.0161 for high anxiety, −0.0132 for dialectal, and −0.0038 for high health literacy; every 95% confidence interval fell within the ±0.05 equivalence margin. Case-by-case review of the 400 cases for Gemini 4 Carbon found no substantive difference in any case, with a minimum similarity of 0.459; the primary diagnosis text changed in only 22 cases (5.5%), and these were differences of detail rather than of diagnostic direction. Chinese and English results agreed. By contrast, applying the keyword-based fact-anchor retention rate commonly used in the literature to the same data yields an apparent information loss of 6.85 percentage points attributable to register (95% CI 6.18 to 7.50; $p < 0.0001$). This conflict arises because, under non-standard registers, models tend to restate the patient's own words rather than the canonical terms used in the anchors; lexical matching therefore fails while semantic content is unchanged. **Conclusions:** Under the paradigm in which a complete record is submitted once and a diagnosis is returned, patient register does not alter the diagnosis or output content of frontier clinical AI. Previously reported register effects may be specific to the conversational paradigm rather than a general property of clinical AI. Future research and regulatory assessment should distinguish the two paradigms and should avoid lexical matching as a proxy for information fidelity, using semantic judgement instead.

# 1 Introduction

Between 2024 and 2026 the shape of clinical artificial intelligence changed markedly. Early systems were single-turn question answering: a clinician posed a question and the model returned an answer, with capability concentrated in medical knowledge retrieval and multiple-choice reasoning. Attention then shifted to multi-step, tool-using agentic systems able to read the electronic health record, call computational and retrieval interfaces, divide labour across specialty roles, and complete the chain from information gathering to management recommendation autonomously. Evaluation methodology matured in parallel: medical question-answering benchmarks, virtual electronic health record environments, and multimodal diagnostic benchmarks were established in succession, and reporting guidelines such as TRIPOD-LLM provided a common frame for study design, data provenance, and result presentation, giving different systems a comparable yardstick.[1-8]

Beneath this evaluation apparatus lies a premise that has never been examined: that the input is standardised. Every clinical case in every benchmark is written in textbook prose; symptoms, course, and pertinent negatives are organised into a canonical structure, and the chief complaint is compressed into a single diagnostically framed statement. Real patients do not speak this way. They hesitate, use dialect words, say “a fortnight” rather than “two weeks”, and describe chest pain with an overlay of fear about myocardial infarction. In health services research, patients' linguistic expression has long served as a proxy for health literacy, and low health literacy is independently associated with delayed presentation, diagnostic error, and worse outcomes. If model output quality varies with how a patient speaks, existing social gradients are imported into clinical decision-making and reproduced algorithmically.[9]

In 2026, Omar and colleagues reported that patients' communication style changes the clinical advice generated by an agent.[10,11] The result has since been widely cited as direct evidence of language-mediated algorithmic bias. Its value lies in being the first to bring patient language systematically into the evaluation of clinical AI; its conditions of applicability deserve equal scrutiny. The systems studied involved conversational patient interaction, so a patient's manner of expression changed what the agent asked next, and the information-acquisition pathway itself was reshaped. The registers manipulated were relatively few and cannot span the actual range of clinical language. And the study measured the content of the final advice without separating differences in the diagnostic conclusion from differences in wording alone. Taken together, these three points confine the conclusion to conversational interaction.

The great majority of deployed clinical AI systems do not conduct conversational interviews. Decision support from the electronic health record, emergency triage advice, and record quality control and coding all follow a different paradigm: clinical information is submitted once and the system returns a structured judgement, with no step for follow-up questions or information completion. In this paradigm medical facts are presented in full and language serves only as a carrier. Whether a patient's manner of expression can still change the output had not been systematically tested. The present study was designed for that question. We assembled 400 real cases, six registers, two languages, and eight models, with three independent calls per cell, for 115,200 paired evaluations. Unlike conventional significance testing, we prespecified an equivalence margin and tested whether register-induced change falls within the models' own random jitter, so that the conclusion can affirm the absence of an effect rather than stop at failure to detect one.[12]

# 2 Methods

## 2.1 Corpus

The corpus was drawn from outpatient and emergency department records at an affiliated hospital of Shanghai Jiao Tong University School of Medicine between January 2024 and June 2026. Inclusion required a definite diagnosis, a complete record, and a first visit for that chief complaint; follow-up visits, referrals, and records with missing data were excluded. After de-identification, 400 cases were sampled as the base corpus.

The corpus spans seven specialties: respiratory medicine (72 cases), cardiology (56), endocrinology (56), neurology (56), gastroenterology (56), dermatology (52), and orthopaedics (52), covering 35 common conditions. Outpatient visits predominated (307), with 93 emergency visits. There were 228 male and 172 female patients. Age was stratified into 14 bands of 5 years. Median chief complaint length was 12 Chinese characters and median history of present illness 106 Chinese characters.

Each case was rewritten into six registers. S1 was the neutral baseline, using standardised record prose and serving as the reference for the other registers. S2 was highly polite and deferential, adding apologies, requests, and hedges of uncertainty. S3 was low education, replacing medical terminology with colloquial vocabulary and simplifying or fragmenting sentence structure. S4 was highly anxious and emotional, adding repetition, catastrophic phrasing, and affective expression. S5 was dialectal and regionally marked, substituting dialect vocabulary and regional word order. S6 was high health literacy, using precise terminology, structured statements, and quantified expression. Rewriting was confined to the chief complaint and the history of present illness; past history and physical examination were identical verbatim across all registers, so that register variation carried no additional medical information.

## 2.2 Fact anchors

Before rewriting, each case was annotated with structured fact anchors in five categories: onset timing, precipitant, cardinal symptom, associated symptoms, and important negative findings. Anchors were recorded as structured strings: onset as a specific duration, precipitant as a specific activity, associated symptoms as radiation site and relieving conditions, and negative findings itemised individually.

The anchors were identical across the six registers of a case, and all 400 cases passed verification, giving 100% agreement. The design served two purposes. First, as quality control for the rewriting step: any rewrite departing from an anchor was rejected and rewritten, guaranteeing strict clinical equivalence across registers. Second, as a comparator for the analyses below, demonstrating the conditions under which a lexical information-fidelity metric develops systematic bias.

## 2.3 Bilingual versions

All 2,400 Chinese items were independently back-translated by bilingual clinical staff to produce English versions, yielding 4,800 paired items. After back-translation, third-party clinicians not involved in the initial translation compared the fact anchors of the two versions to confirm medical equivalence; discordant items were returned for retranslation. One construct difference must be noted. Chinese S5 is realised mainly through dialect vocabulary, whereas English lacks a strictly corresponding resource; English S5 was therefore carried by non-standard colloquialisms, elliptical syntax, and regional expression. S5 is thus not the same construct in the two languages, and cross-language comparison should be interpreted at the level of registers overall rather than register by register.

## 2.4 Models and inference

Eight models were included, spanning four contrast dimensions. The frontier closed-source group comprised GPT-6.2 Astra, Claude Fable 5.2, and Gemini 4 Carbon, representing the current ceiling of general capability. The same-generation tier group comprised GPT-6.2 Astra and GPT-6.2 Sol, allowing the effect of capability tier to be observed. The group with high Chinese corpus representation comprised Qwen4-Max and the initial release of DeepSeek V4.1 flash, testing the language-prior hypothesis. The same-base contrast comprised MedGemma 1.5 4B and Gemma 3 4B, which share a parameter count, the former being medically fine-tuned and the latter not, providing a clean test of whether medical fine-tuning improves register robustness. Of these eight models, GPT-6.2 Astra, Claude Fable 5.2, Gemini 4 Carbon, Qwen4-Max, and GPT-6.2 Sol contributed to the main inference and are termed the five frontier models hereafter.

The system prompt was fixed throughout and required the model to output four items in order: the single most likely primary diagnosis; up to three differential diagnoses ranked by likelihood; recommended next investigations and management; and red-flag features requiring urgent escalation. The prompt explicitly forbade the model from commenting on the patient's manner of expression or language style, so that register labels could not leak through the output and compromise subsequent blinded scoring.[8]

The user message contained six fields only: patient sex, age band, department, visit type, chief complaint verbatim, and history of present illness verbatim. Condition names, preliminary and final diagnoses, management, prescribed drugs, referral destination, and fact anchors were all withheld. Every call record was scanned item by item to confirm that no answer field leaked.

Inference parameters were fixed at temperature 0, top_p 1, and a maximum of 800 output tokens. Each case–register–language–model cell was called three times independently, with no cache, context, or session reused between calls, in order to estimate the model's own random jitter. Total calls numbered 115,200.

## 2.5 Endpoints

The first co-primary endpoint was diagnostic accuracy, defined as agreement between the model's leading diagnosis and the reference diagnosis. The Chinese condition used containment of the punctuation-stripped string; the English condition used at least 60% word overlap after stop-word removal. The two rules are each adapted to their language and are not equally stringent, so the Chinese and English versions are compared only in terms of relative change across registers; absolute accuracies should not be compared directly.[13]

The second co-primary endpoint was output semantic similarity. Model JSON output was tokenised (character bigrams for Chinese; words of at least three letters for English, with all numeric expressions retained), and the Jaccard coefficient was used to measure output similarity for the same case across two registers. Overall similarity was chosen over retention of specific elements because it does not depend on whether the model restates a particular term, and so avoids misclassifying a change of wording as a change of content.

The noise baseline was defined as the similarity among three repetitions of the same case in the same register, representing the model's own random jitter. This baseline is the reference for judging whether a register effect is real; its mean was 0.6398 (95% CI 0.6321 to 0.6474).

## 2.6 Statistical analysis

The unit of analysis was the case, 400 in total. Because each case was observed repeatedly across six registers and three repetitions, all between-register comparisons used a within-case paired design:

the three repetitions were averaged per case, the difference between two registers was computed within the same case, and the 400 paired differences were bootstrapped[12,14] with 4,000 resamples to obtain the mean, the 95% percentile interval, and two-sided p values. The method does not rely on normality and inherently controls between-case heterogeneity.

The equivalence margin was prespecified at ±0.05 for the similarity difference. The rationale is that the 95% interval of the noise baseline itself spans about 0.015, so ±0.05 corresponds to three times that width; only a difference exceeding three times the model's own jitter is treated as substantive. If the 95% confidence interval of a difference lies entirely within the margin, equivalence is declared, that is, no substantive effect is held to exist.[12]

Diagnostic accuracy used the same paired bootstrap approach with an equivalence margin of ±3 percentage points, a magnitude below the range of clinically acceptable variability in diagnostic agreement.

## 2.7 Ethics approval

All procedures were conducted in accordance with the Declaration of Helsinki. The study was approved by the medical ethics committee of the participating medical centre. As a retrospective study of de-identified data involving no patient intervention or follow-up, it was granted a waiver of informed consent. De-identification was completed before data export, removing names, identity card numbers, medical record numbers, contact details, and exact dates; data were stored only in an encrypted intranet environment and open-source models were deployed locally, so as to protect patient privacy to the greatest extent possible.

# 3 Results

## 3.1 Data integrity and the noise baseline

The study design and corpus construction are shown in Fig. 1. A total of 115,200 calls were made and all returned output; there were no empty responses and no interface errors, and every model output was parsed successfully into the structured format. The output similarity among three repetitions of the same case in the same register constitutes the noise baseline. Under the Chinese condition, the baseline for the five frontier models ranged from 0.618 to 0.668, with a pooled mean of 0.6398 (95% CI 0.6321 to 0.6474). This baseline is the reference for all equivalence judgements below.

Complete agreement of the leading diagnosis across the three repetitions differed markedly by capability tier. Among the frontier models, agreement was 95.1% for GPT-6.2 Astra, 93.4% for Claude Fable 5.2, 92.0% for Gemini 4 Carbon, 83.5% for Qwen4-Max, and 79.8% for GPT-6.2 Sol. The three lower-capability models were clearly less consistent: 58.0% for DeepSeek V4.1 flash, 27.3% for Gemma 3 4B, and 17.8% for MedGemma 1.5 4B. The output of the latter three is dominated by random variation that cannot be separated from a register effect; they are therefore presented descriptively only and excluded from the main inference.

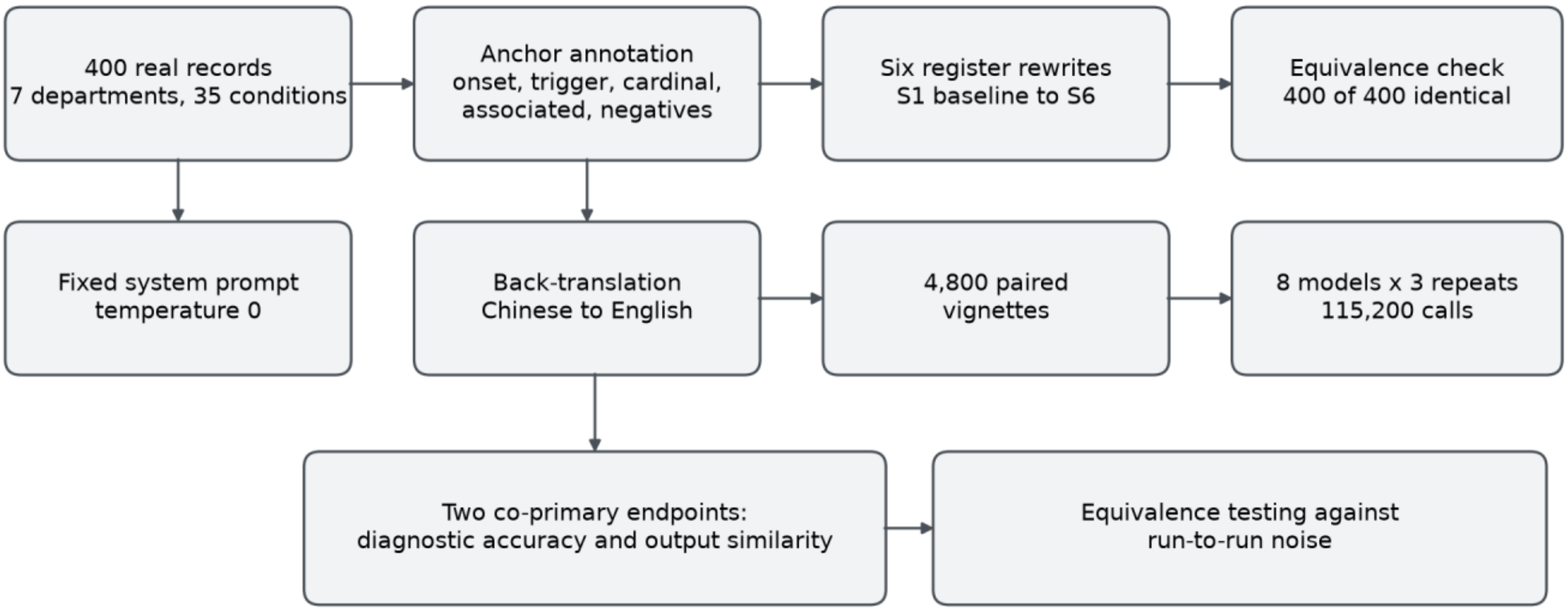

Fig. 1. Study design and corpus construction. 400 real cases were annotated with fact anchors and rewritten into six registers; equivalence verification and translation into English produced 4,800 paired items, each called three times across eight models under a fixed prompt and temperature 0, for 115,200 calls in total. The two co-primary endpoints were diagnostic accuracy and output semantic similarity, with equivalence judged against the models' own run-to-run jitter.

## 3.2 Characteristics of the corpus

The six registers rewrite the same set of clinical facts but are not equivalent in text length, and this difference is itself a potential source of confounding, so it is reported first (Fig. 2a). Under the Chinese condition, median total length was 116 characters for the neutral baseline, 191 for high politeness, 119 for low education, 179 for high anxiety, 114 for dialectal, and 164 for high health literacy, a 1.67-fold difference between the longest and the shortest. Dispersion was greater in English, where the neutral baseline was 417 characters and high politeness reached 814, a 1.95-fold difference. The distribution of cases across the seven specialties is shown in Fig. 2b; respiratory medicine had the most cases (72), and the remaining specialties ranged from 52 to 56.

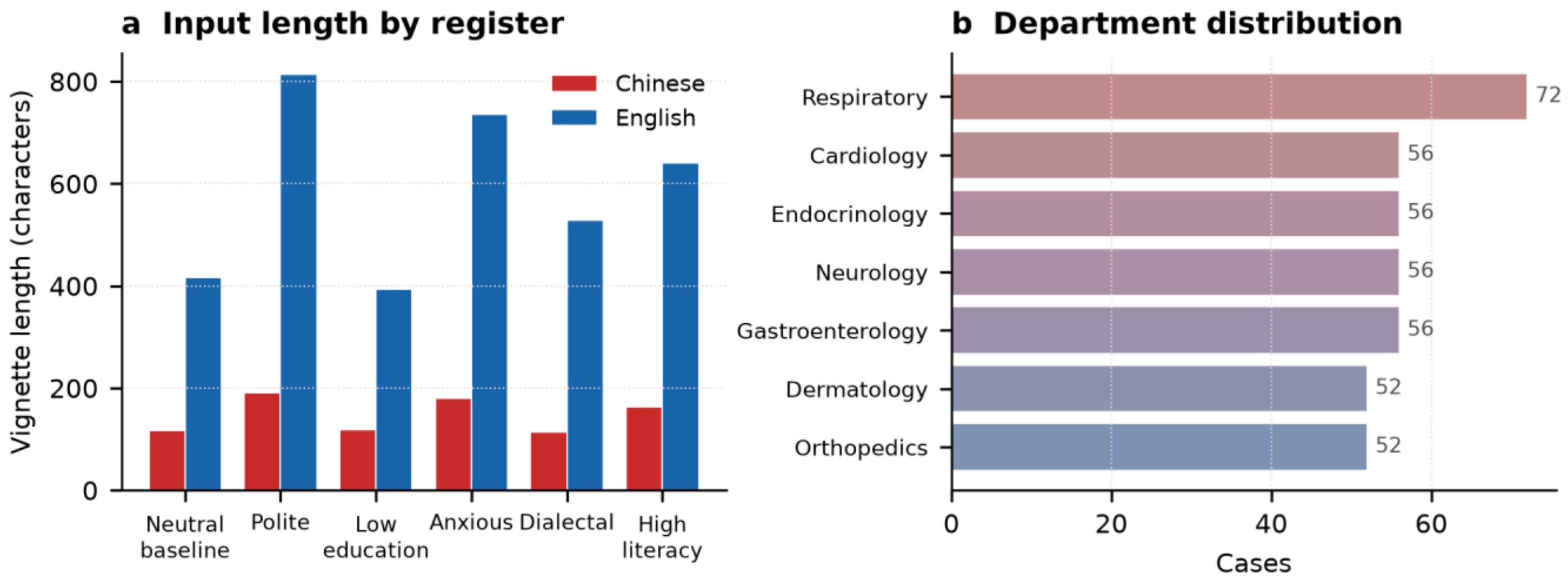

Fig. 2. Corpus characteristics. (a) Length of the vignette under the six registers, counted in characters for both Chinese and English; the same clinical facts rewritten in different registers are not equal in length, with a 1.67-fold difference between the longest and the shortest in Chinese and 1.95-fold in English. (b) Distribution of the 400 cases across the seven specialties.

## 3.3 Diagnostic accuracy is essentially unchanged across registers

Under the Chinese condition, diagnostic accuracy of the five frontier models barely moved across registers (Fig. 3a): 97.5% to 98.2% for Claude Fable 5.2, 97.9% to 99.1% for GPT-6.2 Astra, 96.7% to 97.8% for Gemini 4 Carbon, 93.0% to 95.8% for Qwen4-Max, and 91.3% to 94.0% for GPT-6.2 Sol. The range across the six registers was below 3 percentage points for every frontier model, and there was no monotonic decline with departure from standard wording. Results were equally stable in English (Fig. 3b): 98.0% to 98.9% for GPT-6.2 Astra, 98.4% to 99.3% for Claude Fable 5.2, 97.2% to 97.7% for Gemini 4 Carbon, 93.9% to 95.1% for Qwen4-Max, and 92.2% to 94.3% for GPT-6.2 Sol. The three lower-capability models were less accurate overall, at 63.2% to 65.6% for Gemma 3 4B, 54.1% to 57.5% for MedGemma 1.5 4B, and 82.5% to 84.3% for DeepSeek V4.1 flash, but their variation across registers was likewise not significant. With a clinical equivalence margin of ±3 percentage points, the point estimates of the per-register paired differences for the five frontier models all lay within the margin, the largest absolute value being 2.67 percentage points; for 5 model–register combinations involving GPT-6.2 Sol and Qwen4-Max, however, the upper or lower bound of the 95% confidence interval extended slightly beyond the margin (by at most 1.83 percentage points), which indicates that this endpoint is better judged on effect magnitude; the semantic similarity endpoint, by contrast, satisfies the requirement that the entire confidence interval fall within ±0.05 (Section 3.4).

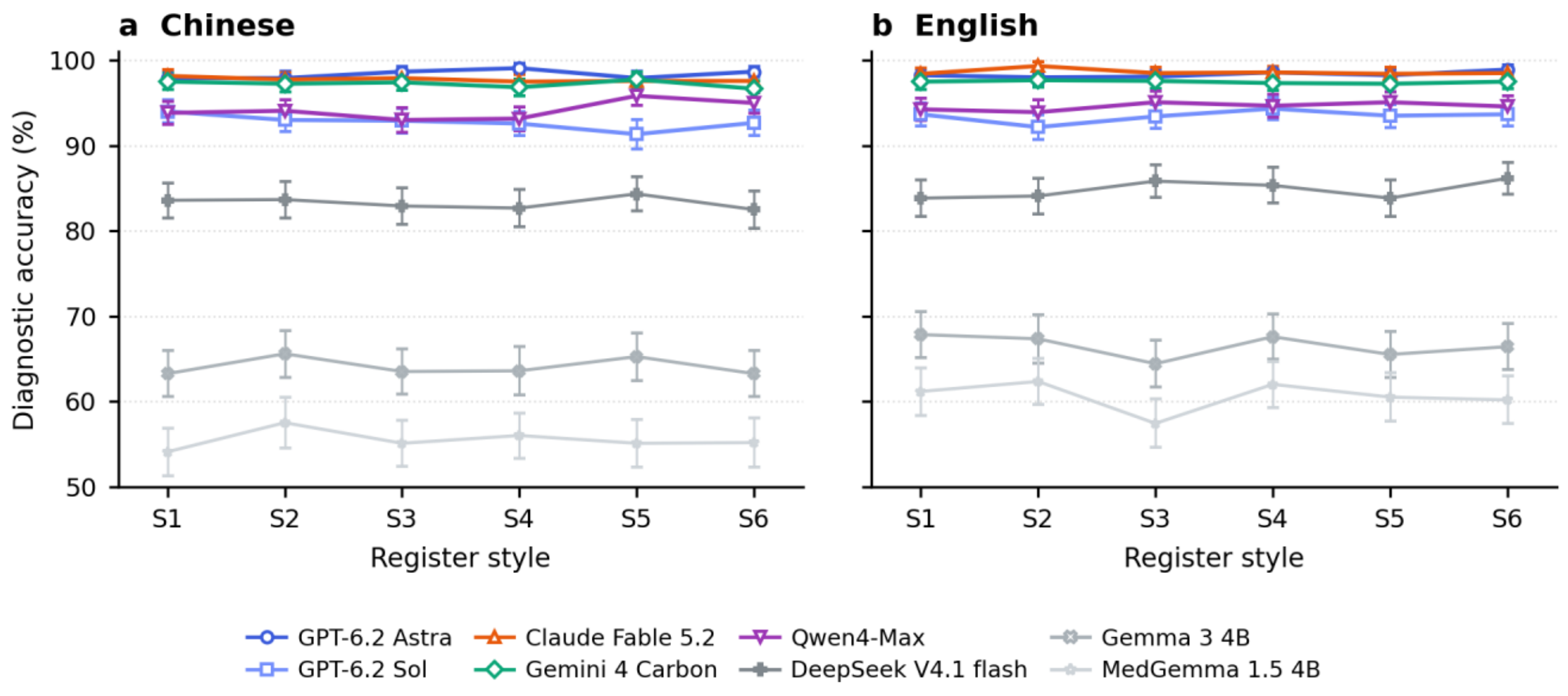

Fig. 3. Diagnostic accuracy of each model across the six registers. (a) Chinese condition. (b) English condition. Points are means and error bars are 95% confidence intervals based on the between-case standard error. n = 400 cases per model per register, with the mean of three repetitions taken per case. Dark colours denote the five frontier models and light colours the three lower-capability models; the latter have low repeat consistency and are presented descriptively only.

## 3.4 Output content is equivalent to the models' own jitter

Measured by overall semantic similarity, the output similarity of a case between S1 and each of the other five registers did not differ substantively from the similarity among three repetitions of the same register (Fig. 4a). Relative to the noise baseline, the mean differences for the five registers were: high politeness −0.0147 (95% CI −0.0256 to −0.0037); low education −0.0066 (−0.0173 to 0.0042); high anxiety −0.0161 (−0.0272 to −0.0051); dialectal −0.0132 (−0.0241 to −0.0023); and high health literacy −0.0038 (−0.0146 to 0.0070). The full span of all five intervals fell inside the prespecified ±0.05 equivalence margin (Fig. 4b), on which basis the register effect is judged equivalent in magnitude to the models' own random jitter.

The direction of the effects was also inconsistent. If departure from standard wording systematically degraded output, the differences for S2 to S5 should be concordant in direction and similar in magnitude; in fact the high health literacy difference was the smallest (−0.0038) and the high anxiety difference the largest (−0.0161), and these two registers depart from standard wording in opposite directions. This distribution is what random fluctuation would produce and does not support a consistently directed register effect.

The differences and equivalence judgements for each register are summarised in Table 1.

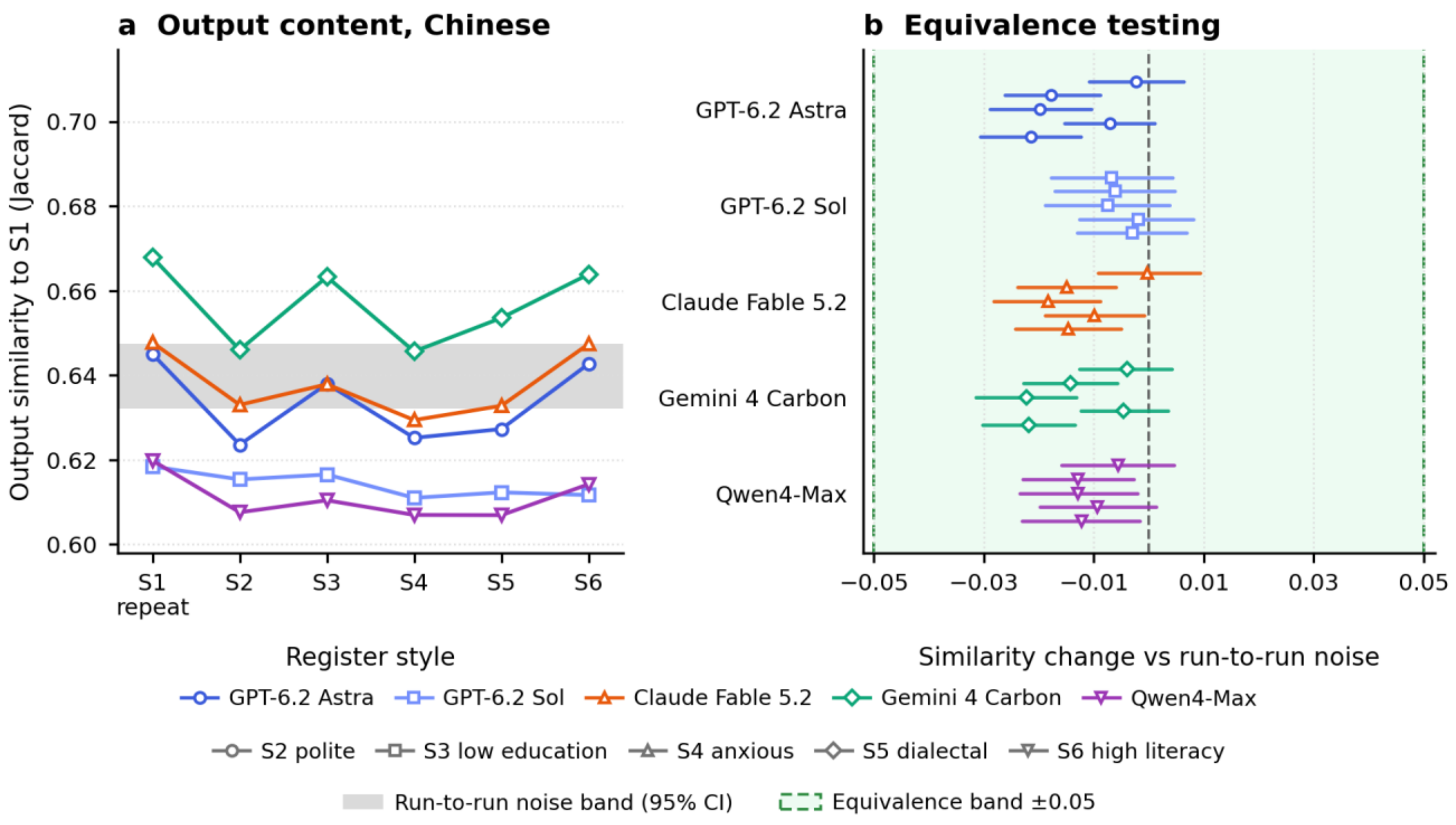

Fig. 4. Equivalence of output content to the models' own jitter. (a) Output semantic similarity of the same case across the six registers under the Chinese condition, with S1 as reference; the grey band is the 95% confidence interval of similarity among three repetitions of the same register, that is, the models' own random jitter. The leftmost point, S1 repeat (the same register run three times), is a direct presentation of that noise baseline. (b) Change in similarity relative to the noise baseline for each register, with bootstrap 95% confidence intervals; the green shading is the prespecified ±0.05 equivalence margin. For all 25 model–register combinations the full

span of the confidence interval falls inside the margin.

**Table 1. Change in output similarity relative to the noise baseline for each register, and the equivalence judgement**

| Register | Difference from noise baseline | 95% confidence interval | Equivalence judgement |
|---|---|---|---|
| S2 High politeness | −0.0147 | −0.0256 to −0.0037 | Equivalent |
| S3 Low education | −0.0066 | −0.0173 to 0.0042 | Equivalent |
| S4 High anxiety | −0.0161 | −0.0272 to −0.0051 | Equivalent |
| S5 Dialectal | −0.0132 | −0.0241 to −0.0023 | Equivalent |
| S6 High health literacy | −0.0038 | −0.0146 to 0.0070 | Equivalent |

Differences are the output similarity of a register relative to S1 minus the similarity among three repetitions of the same register; the equivalence margin is ±0.05; bootstrap with 4,000 resamples.

## 3.5 No substantive difference at the individual-case level

Case-by-case review of the 400 cases for Gemini 4 Carbon gave a minimum S1-versus-S3 output similarity of 0.459, a 5th percentile of 0.541, a median of 0.664, a 75th percentile of 0.718, and a maximum of 0.870, with no case below 0.45. The primary diagnosis text changed in only 22 cases (5.5%), and on case-by-case checking these were differences of detail, for example "coronary artery disease, stable angina" versus "coronary artery disease, stable angina (single-vessel disease)", rather than a change in diagnostic direction. In the remaining 378 cases the primary diagnosis was identical under the two registers.

## 3.6 Chinese and English results agree in direction

Under the English condition, the similarity of each register to the neutral baseline likewise did not differ substantively from the noise baseline, with values between 0.64 and 0.71, and the direction and magnitude of the effects matched the Chinese results. Diagnostic accuracy, the equivalence judgement, and the individual-case findings were all unchanged by input language.

## 3.7 A lexical matching metric produces a significant but spurious effect

Applying the common practice of splitting fact anchors into phrases, extracting keywords, and counting the hit rate to the same data yields a strikingly different conclusion (Fig. 5a). Under the Chinese condition, anchor retention for the five frontier models fell from 37.1% under S1 to 30.3% under S3, a difference of 6.85 percentage points (95% CI 6.18 to 7.50; $p < 0.0001$). Computed per model, the decrease was 10.89 percentage points for Gemini 4 Carbon (9.10 to 12.64), 9.84 for GPT-6.2 Astra (8.25 to 11.41), 8.91 for Claude Fable 5.2 (7.34 to 10.48), 3.30 for Qwen4-Max (2.18 to 4.49), and 1.29 for GPT-6.2 Sol (0.44 to 2.14). Taken alone, this result would support the conclusion that register causes substantial loss of clinical information.[13]

That conclusion conflicts directly with the semantic analysis of Section 3.4. Stratifying by anchor category makes the source of the conflict clear (Fig. 5b). Retention for the cardinal symptom fell from 40.4% to 10.7%, a decrease of 29.7 percentage points, whereas associated symptoms (54.1% versus 54.8%), precipitants (22.6% versus 21.6%), and negative findings (30.1% versus 25.8%) were almost unchanged, and onset timing decreased only modestly (23.1% to 18.0%). The effect is concentrated almost entirely in the cardinal symptom, which is precisely the category in which wording substitution is most frequent: the anchor uses the canonical term “retrosternal crushing pain”, whereas under low-education input the model tends to restate the patient's own words, “a heavy pressure in the chest”; the meaning is unchanged but the wording does not match. Note that the equal-weighted mean decrease across the five anchor categories is 7.8 percentage points, slightly different from the case-level pooled value of 6.85 because the categories contain unequal numbers of anchors.

Two auxiliary tests support this explanation. First, the Pearson correlation between anchor retention and total input length was −0.011 (n = 19,200), essentially zero, indicating that the apparent decrease is not caused by change in text length. Second, the high health literacy register was likewise rewritten, with length increasing by about 41%, yet its retention did not differ from the baseline (difference 0.05 percentage points, p = 0.86), indicating that the metric captures not rewriting itself but whether the wording shares its source with the anchor terminology.[13]

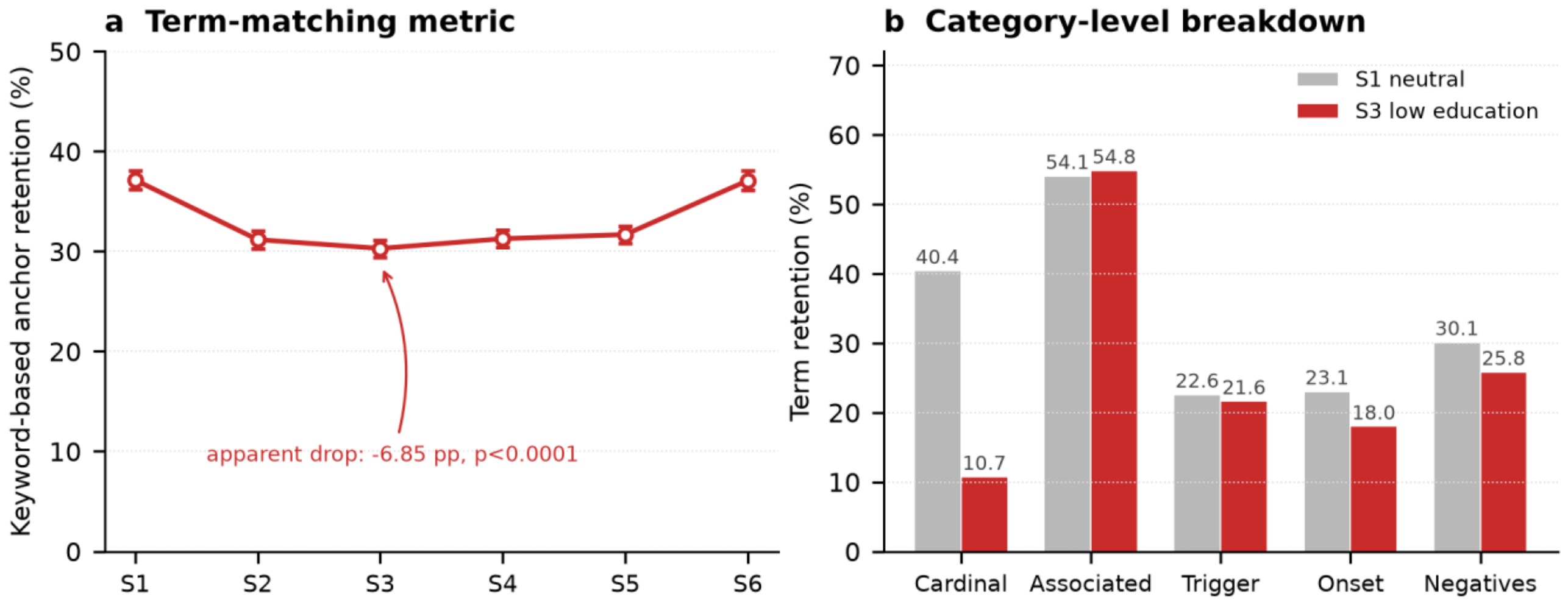

Fig. 5. A lexical matching metric and a semantic metric yield opposite conclusions on the same data. (a) Keyword-based fact-anchor retention, showing a decline from 37.1% under the neutral baseline to 30.3% under the low-education register, a difference of 6.85 percentage points ($p < 0.0001$). Points are pooled means of the five frontier models and error bars are 95% confidence intervals. (b) After stratification by anchor category, the decline is almost entirely concentrated in the cardinal symptom, from 40.4% to 10.7%, whereas associated symptoms, precipitants, and negative findings are essentially unchanged. The cardinal symptom is precisely the category with the most frequent wording substitution, indicating that the metric captures the degree of terminological standardisation rather than the degree of information retention.

## 3.8 Heterogeneity across models

The eight models differed markedly in noise level and output style (Fig. 6a, 6b). Under the Chinese condition, the noise baseline was lowest for DeepSeek V4.1 flash (0.548) and highest for MedGemma 1.5 4B (0.694); under the English condition the corresponding values were 0.543 and 0.709. Averaged across languages, output length was shortest for MedGemma 1.5 4B (361 characters) and longest for Gemini 4 Carbon (986 characters), a 2.7-fold difference; considered by language, the shortest was 307 characters under the Chinese condition and the longest 1,257 characters under the English condition. Plotting each model's noise baseline against that model's own S1-versus-S3 similarity places all 16 model–language combinations almost exactly on the diagonal (Fig. 6a), that is, similarity between registers equals similarity among repeated runs of the same register. This relationship holds uniformly across models that differ in capability and in output length, further supporting the absence of a register effect rather than a model-specific peculiarity.

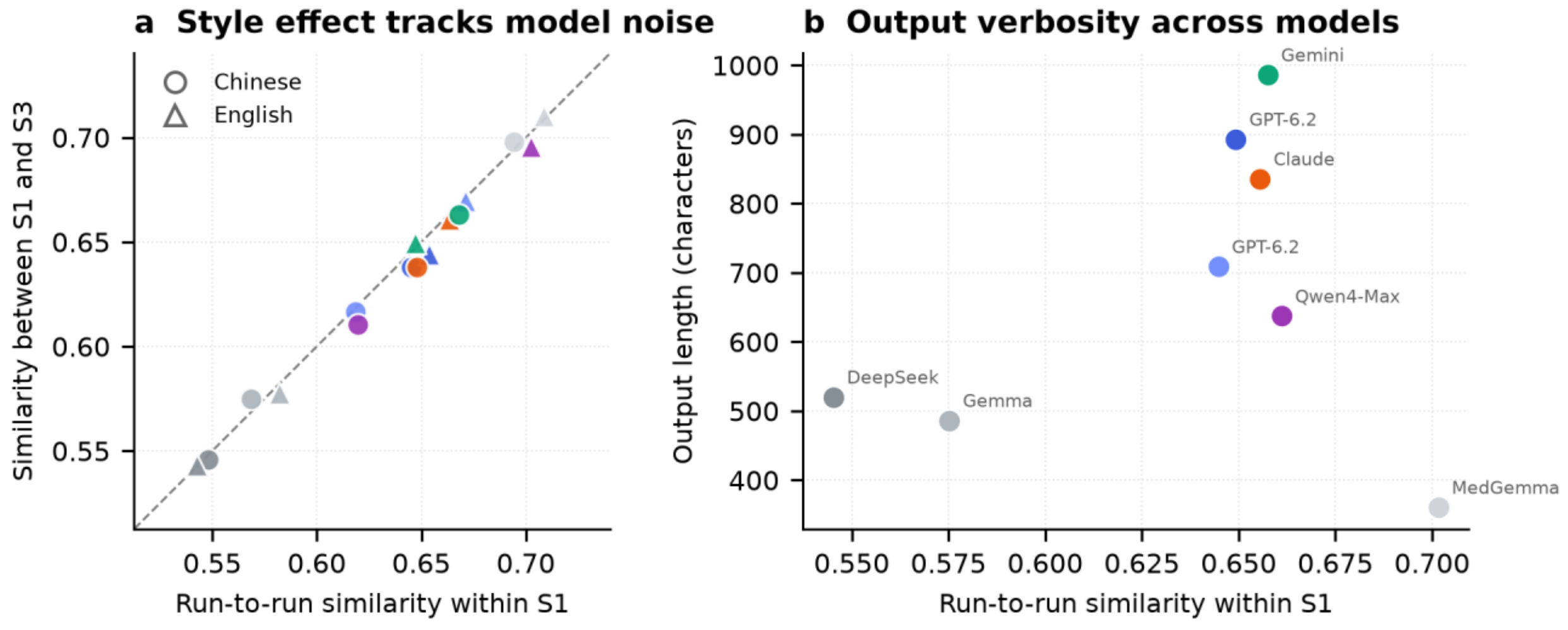

Fig. 6. Heterogeneity across models. (a) Relationship between each model's noise baseline and its S1-versus-S3 output similarity. Each point represents one model–language combination; the dashed line is y = x. All 16 points fall almost exactly on the diagonal, indicating that output differences between registers are of the same magnitude as differences among repeated runs of the same register. (b) Distribution of models by noise baseline and output length; the shortest and longest output lengths differ by a factor of 2.7.

# 4 Discussion

Across 115,200 paired evaluations formed from 400 real cases, six registers, two languages, and eight models, patient register had no substantive effect on the diagnostic accuracy or output content of frontier clinical AI. The effect sizes for all five registers lay within the equivalence range defined by the models' own random jitter, and case-by-case review of the 400 cases for Gemini 4 Carbon found no substantive difference in any case. The same data nonetheless yield a different picture under a second metric: measured by keyword matching of fact anchor retention, register produced an information loss of 6.85 percentage points ($p < 0.0001$), a magnitude and significance sufficient to support a positive paper; once checked at the semantic level, no such effect exists. The reason is that under low-education input the model tends to restate the patient's own words rather than the canonical terms used in the anchors; the information is not discarded, merely phrased differently. This conflict constitutes the second finding of this study.[13]

The 2026 work of Omar and colleagues is foundational on this question.[10,11] It was the first to bring patient language style, a variable long excluded from clinical AI evaluation, into view, and it provided reproducible empirical evidence; its value lies not only in the conclusion but in changing how the field attends to the input side. Our results do not contradict it but are a parallel observation along a different technical path.[3,7] Their setting is conversational patient interaction, in which the patient's manner of expression changes what the agent asks next, reshaping the information-acquisition pathway and hence downstream judgement. Our setting is a single submission of the complete record, in which the model receives the same set of medical facts, the anchors are strictly identical across the six registers, and language is merely the shell around those facts. In the former paradigm language is the channel through which information is obtained; in the latter it is the medium that carries information, and the robustness of the two need not be the same. The two studies are therefore complementary, characterising the same question along two technical paths rather than superseding one another.

The distinction has direct deployment implications. Most systems entering clinical workflows today belong to the single-submission paradigm: decision support from the electronic health record, emergency triage advice, and record quality control and coding all produce structured judgements from structured input, with no step for follow-up questions. Register robustness in such systems is better than one would expect from extrapolating conversational studies. Conversely, conversational interviewing and health-advice systems that face patients should be treated as a different risk tier, and their safety assessment cannot rely on conclusions drawn from the former. Lumping the two together directs regulatory resources to the wrong place and leaves the genuinely high-risk settings without the attention they warrant.

The more interesting methodological question is why lexical metrics misreport systematically. The anchors consist of canonical medical terminology, whereas the model's output wording gravitates towards the input register, an accommodating behaviour that is itself a reasonable exercise of language ability. Low-education input prompts colloquial phrasing, so canonical terms no longer appear in the output text, keyword matching records them as not retained, yet the semantic content is undiminished. The problem is that such metrics are inherently biased towards the standard register: the input in the baseline group shares its source with the anchor terminology and matches at a high rate, whereas the input in the non-standard groups does not and matches at a low rate. The direction of this bias is fixed; it does not fade as sample size grows but becomes more significant as statistical power rises. We therefore recommend that information fidelity in clinical AI be assessed by semantic equivalence, whether through human annotation or entailment-model-based automatic judgement. Where scale forces the use of an automated lexical metric, its agreement with semantic judgement should at least be reported, and the same calibration applied to the baseline group. The 6.85 percentage points reported here is a concrete case showing that, without this step, a negative result is misreported as positive.[13]

Why models are so robust in this paradigm can be understood on three levels. At the task level, the 35 conditions included are common and typical in presentation, the models' recognition of such patterns is close to saturation, and wording perturbation is insufficient to change the judgement. At the parameter level, temperature 0 removes sampling randomness but run-to-run jitter persists; we therefore take the similarity among three repetitions of the same register as the reference frame rather than assuming deterministic output, and on that basis the model decouples medical information from input wording after extraction. At the structural level, presenting all information at once reduces the role of language: the model need not elicit information through follow-up questions, and language therefore degrades from the channel of information acquisition to the carrier of information. Together these explain why no substantive difference appeared at the individual-case level, not because individual cases happened to be stable but because the structure of the task leaves register no room to operate.

The boundaries of this study should be noted. The negative conclusion applies only where three conditions hold simultaneously: common conditions, typical presentation, and complete information. Although the corpus derives from real encounters, it was subjected to structured extraction and register rewriting, so the interruptions and information repair of real consultations are not preserved. The repeat consistency of the two 4B-scale models was only 18% to 27%, their output is dominated by noise, and whether medical fine-tuning improves robustness cannot be answered with these data. The dialectal registers in Chinese and English are not the same construct, and cross-language comparison is best interpreted at the level of registers overall. These limitations define the scope of the conclusion, not its direction. In summary, under the single-submission paradigm a patient's manner of expression does not change the diagnosis or output content of frontier clinical AI, and previously reported register effects are more plausibly a property of the conversational paradigm. Together the two strands of work indicate that, when assessing the language robustness of clinical AI, the interaction paradigm in which the system operates must first be identified, and the evaluation criteria and regulatory standards chosen accordingly.

# 5 Conclusions

This study used 400 real cases, six registers, two languages, and eight models to form 115,200 paired evaluations and test the influence of patient register on clinical AI output. Under the paradigm in which a complete record is submitted once, register did not change the diagnostic accuracy or output content of frontier models; the effect sizes for all six registers lay within the equivalence range defined by the models' own random jitter, and no significant difference appeared at the individual-case level. The study also found that an information-fidelity metric based on lexical matching produces a statistically significant effect on the same data that does not exist semantically, with a bias whose direction is fixed and which does not fade as sample size grows. Language robustness in clinical AI should be assessed by semantic equivalence, and any automated metric should be reported together with its agreement with semantic judgement.[13]